\def\year{2020}\relax
\documentclass[letterpaper]{article} 
\usepackage{aaai20}  
\usepackage{times}  
\usepackage{helvet} 
\usepackage{courier}  
\usepackage[hyphens]{url}  
\usepackage{graphicx} 
\usepackage{graphicx}  
\usepackage{amsmath}
\usepackage{algorithm}
\usepackage[noend]{algpseudocode}
\usepackage{graphicx}
\usepackage{subcaption}
\usepackage{color, colortbl, multirow}
\usepackage{amssymb}
\usepackage{comment}

\title{From Time Series to Euclidean Spaces: On Spatial Transformations for Temporal Clustering}

 \author{Nuno Mota Goncalves \\
IBM Research - Zurich \\
glv@zurich.ibm.com
\And Ioana Giurgiu\\
IBM Research - Zurich\\
igi@zurich.ibm.com
\And Anika Schumann\\
IBM Research - Zurich\\
ikh@zurich.ibm.com}

\begin{document}

\maketitle

\begin{abstract}
Unsupervized clustering of temporal data is both challenging and crucial in machine learning. In this paper, we show that neither traditional clustering methods, time series specific or even deep learning-based alternatives generalize well when both varying sampling rates and high dimensionality are present in the input data. We propose a novel approach to temporal clustering, in which we (1) transform the input time series into a distance-based projected representation by using similarity measures suitable for dealing with temporal data, (2) feed these projections into a multi-layer CNN-GRU autoencoder to generate meaningful domain-aware latent representations, which ultimately (3) allow for a natural separation of clusters beneficial for most important traditional clustering algorithms. We evaluate our approach on time series datasets from various domains and show that it not only outperforms existing methods in all cases, by up to 32\%, but is also robust and incurs negligible computation overheads.
\end{abstract}

\section{Introduction} \label{intro}
One major concern in machine learning is how to handle data with temporal characteristics - or \textit{time series}. Such data exists in staggering amounts in almost every field, from sensors, social networks, IoT, human mobility, finance or even medicine~\cite{review1,review2}. Unlike static data, there exist strong temporal and spatial dependencies (e.g., correlated time series), so an appropriate treatment of these properties becomes critical in any temporal data processing. Among all techniques applied to time series, \textit{unsupervised clustering} is the most widely used for two main reasons. First, it does not require costly supervision or time-consuming annotation of data~\cite{kshape}. Second, it not only is a powerful stand-alone exploratory method but also proved to be a useful pre-processing or post-processing step for other tasks (e.g., anomaly detection). 

In the literature, various approaches for clustering temporal data have been proposed. When the time series have the same length and are sampled at the same frequency, methods like k-means, which employ the Euclidean distance, can be used~\cite{review,charu}. However, most times the problem of time series clustering remains challenging as typically temporal data originating from different domains shows significant variations in dimensionality and temporal scales. To tackle different sampling rates, elastic distance metrics such as dynamic time warping (DTW)~\cite{dtw} or cross-correlation metrics such as SBD, used in k-shape~\cite{kshape}, are frequently used. However, these metrics suffer from high-dimensionality in univariate spaces (e.g., long temporal sequences, where either the computation costs become intractable or an optimal alignment becomes non-trivial and possibly ambiguous). This effect is especially exacerbated for multivariate time series. 

In other domains, dimensionality reduction performed by autoencoders (AEs) has allowed for several improvements in unsupervised classification~\cite{dec,idec,dcc,dcn,mixae,jule,depict,damic}. The idea is to generate lower-dimensional latent representations that are then fed into a customized layer responsible for the clustering task (i.e., typically by minimizing clustering loss via Kullback-Leibler (KL) divergence). When applied to image and text benchmarks, these approaches significantly outperform traditional methods. However, despite their achievements, they still lack a principled cross-domain generalization. First, they depend on standard target distributions (required by KL divergence) which requires tuning for each data type. Second, they have an implicit expectation as to the latent spaces' descriptiveness of the target domain (i.e., samples that should be clustered together would be closely represented in the latent space). This may not hold if models capture undesirable properties of the data (e.g., artifacts) or in spaces where spacial proximity is a sub-optimal measure of sample similarity (e.g., the time domain). 

In this paper we postulate that all existing approaches suffer from significant drawbacks when applied in time series clustering, since: (1) traditional clustering methods cannot deal well with varying sampling rates and lengths; (2) domain-specific clustering approaches greatly suffer from high dimensionality in the sample space; and (3) AE-based methods have no guarantees to generate meaningful latent representations. Instead, we propose a deep temporal clustering approach that focuses exclusively on generating latent spaces that naturally separate clusters, while at the same time preserving domain-specific characteristics of the data. We achieve this by: 
\begin{itemize}
     \item feeding the input space into a multi-layer CNN-GRU AE, in order to extract temporal waveforms from the original time series in a compressed format;
     \item before computing the reconstruction error, applying spatial transformations to both the reference samples and their reconstructed counterparts to guide the latent space construction;
     \item using the embeddings into a final clustering step, where any traditional clustering method could be used (e.g., k-means, spectral, DBSCAN). 
\end{itemize}

In this paper, we demonstrate that by combining spatial transformations with a multi-layer CNN-GRU AE, we can generalize Euclidean metric-based clustering algorithms to potentially non-euclidean domains. We evaluate our approach on time series datasets from various sources and show that it outperforms existing methods in all cases, both in terms of accuracy (up to an 32\% boost) and robustness in parameterization - while incurring small or negligible computation overheads.

\section{Related Work}
\textbf{Distance metrics-based clustering} -- Measuring the distance between different time series needs to take into consideration the temporal correlation between the data points in a time series and the complex nature of the noise that may be present. Such noise can, for instance, be represented by different sampling rates~\cite{sampling}. In the simplest case, when the time series have the same length and are sampled at the same frequency, general clustering methods based on Euclidean distance, such as k-means, can be used~\cite{review}. However, these do not perform well when sampling rates between comparing time series are different. In this case, elastic distance measures such as dynamic time warping (DTW)~\cite{dtw}, edit distance on real sequence (EDR)~\cite{edr} or edit distance with real penalty (ERP)~\cite{erp}, as well as cross-correlation measures such as the SBD metric used in k-shape~\cite{kshape} are popular choices. These measures mostly focus on comparing the shape of the time series, but a recent review of most popular time series distances found that none of them is more robust than the others to all the different kinds of noise commonly present in temporal data~\cite{trajreview}.

\textbf{DNN-based clustering} -- Deep embedding clustering approaches~\cite{review} have recently been proposed to deal with high-dimensional data, in particular images and text. For instance, DEC~\cite{dec} uses stacked denoising AEs to generate latent representations. However, spatial relationships in the reconstructed samples cannot be guaranteed. To alleviate this drawback, IDEC~\cite{idec} defines a joint optimization objective that minimizes KL divergence simultaneously with the reconstruction loss. Similar to IDEC, DCC~\cite{dcc} and DCN~\cite{dcn} use the reconstruction loss as part of the optimization objective. In DCC, the authors assume the number of clusters is unknown in advance and propose a clustering algorithm that continuously performs the clustering task by optimizing an objective that does not need to be updated during the optimization phase. Differently, DEPICT~\cite{depict} uses discrete reconfigurations of datapoints to centroids and in ~\cite{jule} the authors use agglomerative clustering. DCN uses a novel alternating stochastic gradient algorithm and a cluster structure-promoting regularization to achieve superior accuracy compared to IDEC. DAMIC~\cite{damic} proposes an algorithm based on mixture-of-experts, where each cluster is represented by an AE. 
Similarly, MIXAE~\cite{mixae} proposes to train simultaneously a set of AEs via a composite objective function, jointly motivating low reconstruction error and cluster identification error.


In the time series domain, deep neural networks have primarily been used for supervized learning, with few exceptions~\cite{timenet,survey}. 
However, to the best of our knowledge, there have been no efforts proposing deep learning approaches for time series clustering that are invariant to affine transformations.

\section{Approach}\label{approach}

Our approach is motivated by the following observations:
\begin{itemize}
    \item traditional clustering methods, such as k-means, DBSCAN or spectral clustering are not well-suited for temporal data where varying sampling rates and sample lengths will frequently occur;
    \item time series specific clustering methods, such as k-means (using DTW) and k-shape, will greatly suffer from high-dimensional samples (i.e., length of time series) where an optimal alignment becomes non-trivial and possibly ambiguous;
    \item common deep learning approaches for dimensionality reduction, such as AEs, have no guarantees as to the meaningfulness of their latent representations, either degrading clustering accuracy for complex domains~\cite{autowarp} or making it unpredictable.
\end{itemize}

In this paper we take a bottom-up approach. We start by enabling domain-aware latent representations using AEs. Then, we leverage well-known temporal similarity measures and their domain-specific properties. As a last step, we propose a pipeline to better generalize traditional clustering algorithms to other (potentially non-euclidean) domains.

\subsection{Meaningful Latent Spaces}


Consider an unlabeled dataset $S$ with \textit{N} samples, where $S_{i} \in R^{V \times T_{i}}$ $\forall \{i \in \mathbb{N} \bigm| i \leq N \}$ ($V$ being the number of variables and $T_{i}$ the number of time steps specific to $S_{i}$). Generally, learning a latent representation for each input sample can easily be achieved through non-linear mappings performed by an autoencoder. These mappings can be defined as $f_{W} : S_{i} \rightarrow z_{i}$ and $g_{W} : z_{i} \rightarrow S_{i}^{'}$, where $z_{i}$ is the latent space embedding of $S_{i}$ and $S_{i}^{'}$ is the reconstruction of $S_{i}$. $f_{W}$ and $g_{W}$ will be learnt through a sample reconstruction error, $L_{r}$, as follows:

\begin{equation}
    L_{r}(S_{i}, S_{i}^{'}) = \frac{1}{\left|S_{i}\right|}\sum^{\left|S_{i}\right|}_{j} \left(S_{ij} - S^{'}_{ij}\right)^{2}
\nonumber
\end{equation}

This error is nothing more than the mean squared error (MSE) between two samples. The problem with this measure is that it is analogous to Euclidean distance, arguably not the best metric for comparing time series~\cite{charu}. Not only would this measure not be ideal to quantify reconstruction quality in a sample-by-sample basis, but AEs implicitly enforce that the whole reconstructed space is similar to the original space. Even if this is a desired property in other scenarios, our ultimate goal is clustering in the latent space. Therefore, our focus shifts from ensuring that the AE achieves low reconstruction error to providing relevant properties for traditional clustering, such as invariance towards geometric transformations in the input space.

To tackle the aforementioned limitations, we propose a transformation $f$ that represents a sample as its \textit{similarity} to each one of $p$ randomly picked \textit{pivot points}, detailed in lines 1--6 of Alg.~\ref{algo}. As such, applying $f$ to a sample $S_{i} \in R^{V \times T_{i}}$ generates a new sample $f(S_{i}) \in R^{p \times W}$, where $W$ is the dimensionality of the used \textit{similarity} function (e.g., when comparing two multivariate time series, it may be useful to calculate the pairwise similarity between each variable rather than enforcing a single value to compare the whole sample). Finally, we redefine $L_{r}(S_{i}, S_{i}^{'})$ as:

\begin{equation}
    L_{r}(S_{i}, S_{i}^{'}) = \frac{1}{p}\sum^{p}_{j} \left(f(S_{i})_{j} - f(S^{'}_{i})_{j}\right)^{2}
\nonumber
\end{equation}

\noindent where f($S_{i}$) and f($S^{'}_{i}$) are transformed representations of $S_{i}$ and $S^{'}_{i}$. Intuitively, this new error will be minimized if samples remain \textit{relatively similar} to the pivot points in the reconstructed space, even if the space as a whole has changed.

To better illustrate our point, we will define our domain's \textit{similarity} with the Euclidean distance. While we do not employ it in the time series domain, we will use it in the following explanations due to its intuitive value. Thus, we formally define a \textit{rigid transformation} as a transformation that satisfies $Euc(L(X),L(Y))^2 = Euc(X,Y)^2$, where $L : R^n \rightarrow R^n$. In this context, $L$ simply represents a linear transformation of a vector space. Since our transformed space will be solely composed of Euclidean distances, it will then be invariant to any \textit{rigid transformation} applied to our initial space.

As an example, we can prove that a \textit{translation} is a \textit{rigid transformation}. By defining $L(w) : w \rightarrow w + v$ (a translation given by a vector $v$):

\vspace{-1ex}
\begin{equation}
\begin{split}
    Euc(L(X),L(Y))^{2} &= Euc(X+v, Y+v)^{2} \\
    &= (X + v - Y - v) \cdot (X + v - Y - v) \\
    &= (X - Y) \cdot (X - Y) \\
    &= Euc(X,Y)^{2}
\end{split}
\nonumber
\end{equation}

\begin{figure}
\centering
\includegraphics[width=0.95\linewidth]{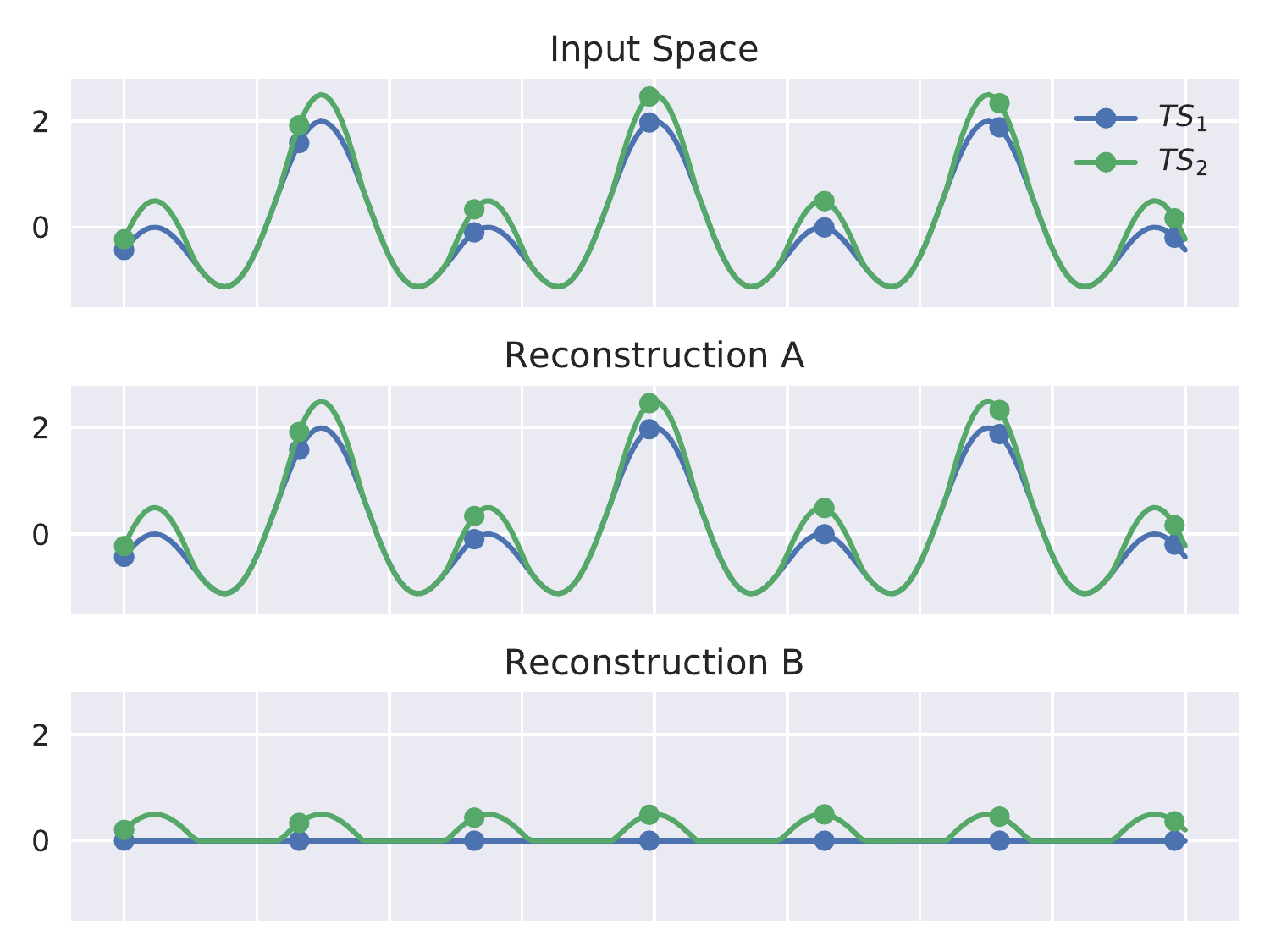}
\caption{Alternative reconstructions of the input space. \textit{Reconstruction A} would be the ideal reconstruction through a common AE, whereas \textit{Reconstruction B} is a reconstruction allowed through our transformations. Both \textit{Reconstruction A} and \textit{Reconstruction B} have exactly the same $L_{r}$, with \textit{Reconstruction B}'s representation being a rigid transformation of \textit{Reconstruction A} ($h(w) : w \rightarrow w - v$, where $v = TS_1$).}
\label{fig:alternative_reconstructions}
\end{figure}

By defining $L$ accordingly, the same logic would immediately follow for rotations and, if the transformed space is normalized, we can also extend this property to uniform scaling. These transformations should not be confused with actually transforming the visual representations of our samples. A time series with $n$ timestamps will be considered an $n$-dimensional vector and any transformation can be applied to any of its dimensions (timestamps) independently of the known temporal dependence. Fig.~\ref{fig:alternative_reconstructions} illustrates this effect with a translation, and not only shows that our model can learn simplified representations of the input data, but also that its only constraint is to preserve \textit{dissimilarities} between samples in the input space. The amount of \textit{filtering} our model is allowed to perform will be mostly constrained by its complexity and latent space dimensionality. The simpler the model, the simpler the reconstructed space. This is true for any AE, but usually at the cost of an ever-increasing $L_{r}$ as the models stop being able to recreate all the nuances of the initial samples. Since we allow simpler representations to minimize $L_{r}$, we can achieve this with minimal penalty to the latent space quality (and potentially no penalty to $L_{r}$).

\algnewcommand\algorithmicforeach{\textbf{for each}}
\algdef{S}[FOR]{ForEach}[1]{\algorithmicforeach\ #1\ \algorithmicdo}

\begin{algorithm}[t]
\caption{Get clusters from embeddings obtained from $projected$ space}\label{algo}
\begin{algorithmic}[1]
\Procedure{GenProjSpace}{$S_{i=1}^{N}$, $P_{j=1}^{p}$ $\subset$ $S$}
\State $S^{'}$ = $matrix_{N \times p}$ 
\ForEach {[sample, i] in $S$}
\ForEach {[pivot, j] in $P$}
\State $S^{'}$[i, j] = Dist(sample, pivot)
\EndFor
\EndFor
\State \Return $S^{'}$
\EndProcedure
\Procedure{GetClusters}{$S_{i=1}^{N}$, $P_{j=1}^{p}$ $\subset$ $S$, $f_{W}$, $k$ clusters}
\State $S^{'}$ = $GenProjSpace(S, P)$ 
\State $z$ = $f_{W}(S^{'})$
\State $clusters$ = kmeans($z$, $k$)
\State \Return $clusters$
\EndProcedure
\end{algorithmic}
\end{algorithm}

\subsection{Leveraging Temporal Distance Metrics}



Sound alternatives to the Euclidean distance for temporal data are DTW or SBD. While DTW is extremely well-known and popular for comparing time series, its complexity is quadratic on the dimensionality of the samples. There exist optimizations to the algorithm but, just like the original implementation, require parameter tuning for optimal results -- a significant constraint in a completely unsupervised setting. On the contrary, SBD is specifically built for temporal data k-Shape~\cite{kshape} and does not require any parameterization. It also generally provides better results than DTW and its variants, leading us to believe it is the best metric to report our results on. SBD is based on cross-correlation and uses coefficient normalization (i.e., between -1 and 1), independent of data normalization. The coefficient normalization divides the cross-correlation sequence by the geometric mean of autocorrelations of the individual sequences, as below: 

\vspace{-1ex}
\begin{equation}
SBD(\vec{x}, \vec{y}) = 1 - \underset{w} max(\frac{CC_{w}(\vec{x}, \vec{y})}{\sqrt{R_{0}(\vec{x}, \vec{x}) R_{0}(\vec{y}, \vec{y})}}) \nonumber
\end{equation}
\vspace{-1ex}

\noindent where $w$ is the position where the coefficient normalization is maximized, $CC_{w}(\cdot, \cdot)$ is the cross-correlation sequence with length 2m-1 (i.e., m = length of $z_{i}$) and $R_{0}(\vec{x}, \vec{x})$ = $\sum_{l=1}^{m} x_{l} x_{l}$. SBD ranges from 0 to 2, with 0 representing perfect similarity between two sequences. Based on the definition of $CC_{w}(\cdot, \cdot)$, its temporal complexity is quadratic but, by using FFT, it can be reduced to $\mathcal{O}$($m$ log($m$)).

Due to the more complex nature of this measure, exploring the types of geometric transformations now allowed by our transformations would require a thorough analysis of their own, and will hence be left for future work. Regardless, it is equally easy to describe the invariance properties associated with the new generated transformations (explained in-depth in k-Shape): (1) scaling and translation; (2) shift; (3) uniform scaling; (3) occlusion; and (4) complexity. Although being invariant to different properties, the main premise of our transformation remains intact: a lower $L_{r}$ will only be achieved by spaces in which samples remain \textit{relatively similar} to the chosen pivot points (the only difference being in how \textit{similarity} is calculated).

\subsection{Tying It All Together}\label{approach-tying-it-all}


Drawing from our previously learned encoder mapping, $f_{W}$, the embeddings $z_{i}$ of our AE are then fed into a classic clustering algorithm to group $S$ into $k$ clusters. This approach differs from other literature on unsupervised clustering by not relying on a KL divergence-based approach and instead generating latent spaces that preserve desirable domain-specific properties. We note the following: (1) using KL divergence requires defining a target probability distribution that is specific to the domain, which has little generalization properties; and (2) any traditional clustering algorithm can still be used at this step (e.g., k-means) as the generated embeddings are already representative of the domain, clearly separating clusters as shown in Fig.~\ref{fig:latent}. Even if our approach still requires us to pick a similarity metric, due to the nature of SBD this is a much smaller constraint because it requires no parameterization tuning. Nonetheless, other works have proposed methods for automatically deriving warping distances for a given input space~\cite{autowarp}. This could allow us to further improve our transformations' performance while remaining in a fully unsupervised setting but, given the complexity of this method alone, it will be left for future work.

\begin{figure*}[t]
\centering
  \includegraphics[width=0.95\linewidth]{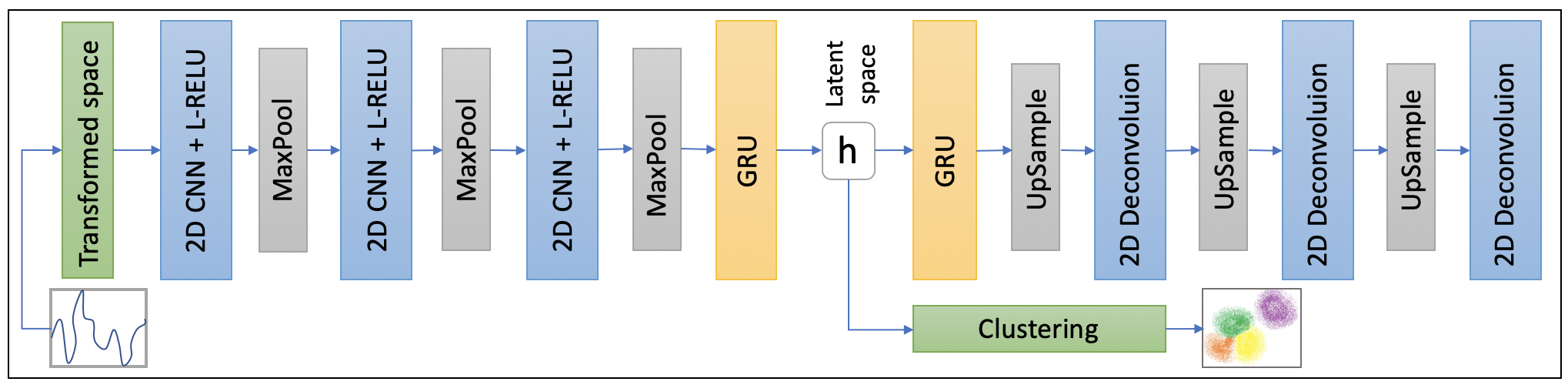}
  \vspace{-1ex}
  \caption{Schematic architecture of our approach. Spatially transformed inputs are processed through a multi-layer CNN-GRU symmetric AE. The encoder produces a latent representation $h$, which is then fed to a clustering layer.} 
  \label{architecture}
\end{figure*}

In order to avoid generating transformations every time we calculate $L_{r}$ during training, we apply these transformations directly on the input space, that is before even feeding them to the AE. Since $f(S)$ is already a simplified representation of our samples, it should allow for faster training and more robust results, but the optimization will only yield similar results if and only if $\frac{1}{N}\sum_{i}^{N} L_{r}\left(f(S_i), f(S_i^{'})\right) \Leftrightarrow \frac{1}{N}\sum_{i}^{N} L_{r}\left(f(S_i), f(S_i)^{'}\right)$. This means that the loss associated with transforming the reconstruction has to be equivalent to the loss of reconstructing the transformations. This should intuitively be true if our model is complex enough to accurately learn both $S$ and $f(S)$ -- driving our choice in the AE's architecture (Fig.~\ref{architecture}). The main inconvenience of this optimization is that we can no longer visualize the alternative representations learned by our AE, unless the assigned similarity metric is bijective (allowing full sample reconstruction from distances alone). Even if, for instance, we could achieve such a property by (i) defining Euclidean distance as our similarity metric and (ii) using $d+1$ pivot points for $d$-dimensional samples, bijection is not a necessity. Truthfully, by using either DTW or SBD it might not even be a possibility (i.e., shown for DTW in~\cite{bijection}). Through our results, however, we show that not only can we ignore this property but also greatly reduce the number of required \textit{pivot points} without sacrificing accuracy. The final clustering pipeline can be found in lines 7--11 in Alg.~\ref{algo}.

Finally, and to expand on the AE's architecture, the first three levels consist of 2D convolution layers, which extract key short-term features, each followed by a max pooling layer of size P. The goal is to cast the time series into a more compact representation while retaining the relevant information. This dimensionality reduction is crucial for further processing to avoid very long sequences which can lead to poor performance. Leaky rectifying linear units (L-ReLU) are used. These activations are then fed to a gated recurrent unit (GRU) to obtain the latent representation. Using recurrent neural networks in the AE is inspired by their successful application to sentiment classification~\cite{seqlearning} and machine translation~\cite{translation}, while the use of convolutional networks has been shown to be effective when dealing with temporal data in~\cite{nin}. This allows for the input to be collapsed in all dimensions except temporal. More specifically, the embeddings capture overarching features of the time series, while learning to ignore outliers and sampling rates, as shown in~\cite{autowarp}. Reconstruction is provided by a GRU layer, followed by three upsampling layers of size P interleaved with three deconvolutional layers.

\begin{table*}[t]
\centering
\begin{tabular}{ |c|c|c|c|c|c|c|c|c|c| } 
\hline
&\textbf{Model} & \textbf{A} & \textbf{B} & \textbf{C} & \textbf{D} & \textbf{E} & \textbf{F} & \textbf{G} & \textbf{H} \\\hline 
\multirow{5}{*}{OS} &k-means+EUC & 51.6$\pm$5 & 61.6$\pm$4&51.5$\pm$2 &55.2$\pm$2&83.3$\pm$7&53$\pm$3&64.6$\pm$10 &35.5$\pm$5\\
 &k-means+DTW & 55.8$\pm$6& 64.1$\pm$5&50.5$\pm$2&55$\pm$2&83.8$\pm$7&63$\pm$3&55$\pm$11 &41.4$\pm$6\\
 &k-shape& 52.6$\pm$6 & 58.2$\pm$3&54.7$\pm$14&52.7$\pm$1&64.7$\pm$7&51$\pm$1&49$\pm$4 &41$\pm$8\\
 &spectral & 51.6$\pm$5 &59$\pm$4&52.6$\pm$5 &54.6$\pm$1&82.8$\pm$8&51.5$\pm$2&40$\pm$5 &43$\pm$4\\
 &dbscan & 50$\pm$4 & 44.7$\pm$5&58.2$\pm$10&49.7$\pm$2&18$\pm$3&50$\pm$2&16.6$\pm$4 &31$\pm$7\\\hline
\multirow{3}{*}{LS} &k-means+EUC & 52$\pm$4&55$\pm$4&64$\pm$4&55$\pm$1&83.3$\pm$2&52$\pm$1&57$\pm$5 &60$\pm$4\\
 &spectral & 52.4$\pm$3 & 61$\pm$4&59.3$\pm$6&54.5$\pm$1&83.3$\pm$2&54$\pm$2&41$\pm$6 &55$\pm$4\\
 &dbscan& 34$\pm$4 &59.3$\pm$3&48$\pm$8&60.7$\pm$2&85.7$\pm$1&50$\pm$2&16.6$\pm$4 &47$\pm$5\\\hline
\multirow{3}{*}{Pr} &k-means+EUC & 62.8$\pm$4 &61.1$\pm$5&71.4$\pm$1&60.6$\pm$1&87$\pm$2&81.5$\pm$2&79$\pm$7&65$\pm$3\\
&spectral& 56.4$\pm$3 &59.3$\pm$4&54.5$\pm$2&64$\pm$1&92.3$\pm$3&83.5$\pm$1&67$\pm$8&63$\pm$4\\
 &dbscan& 35.2$\pm$3 &65$\pm$4&53$\pm$3&64.5$\pm$2&77$\pm$2&50$\pm$1&17$\pm$5 &55$\pm$3\\
 
 & \textbf{Impr.} [OS/LS] [k-means] &7 &-3 &7.4 & 5.4&3.2 &18.5 &14.4& 5\\
& \textbf{Impr.} [OS/LS] [spectral] &4 &-1.7 &-4.8 & 9.4&9 &29.5 & 26& 8\\\hline
\multirow{4}{*}{Pr+LS} &k-means+EUC & \textbf{67.6}$\pm$2 &70$\pm$2&\textbf{78}$\pm$1 &\textbf{66.7}$\pm$1&\textbf{94.4}$\pm$1&\textbf{85.5}$\pm$1&\textbf{84}$\pm$5 &78$\pm$2\\
&spectral & 66.4$\pm$2 & \textbf{72.8}$\pm$3 &64$\pm$3&65.1$\pm$1&93.8$\pm$2&84.7$\pm$1&73$\pm$5 &\textbf{82}$\pm$2\\
 &dbscan  & 41.2$\pm$1 &70.2$\pm$2&61$\pm$2&65.2$\pm$2&87$\pm$2&55$\pm$2&23$\pm$4 &72$\pm$2\\
& \textbf{Impr.} [OS/LS] [k-means] & 11.8 & 5.9 & 14 & 11.5& 10.6 &22.5 & 19.4& 18
\\
& \textbf{Impr.} [OS/LS] [spectral] &14 &11.8 &4.7& 10.5&10.5 &30.7 & 32& 27
\\
\hline
\end{tabular}
\caption{Accuracy (mean and standard deviation over 10 runs) for our approach (\texttt{Pr+LS}) and \texttt{Pr}, compared to two baselines: \texttt{OS} and \texttt{LS}, when 16 pivot points are used to generate the projected space. For each dataset, we indicate $k$ = number of clusters, $n$ = number of samples and $t$ = time series length, as follows: \textbf{A} = Computers(2,500,720), \textbf{B} = ECG5000(5,5000,140), \textbf{C} = ItalyPowerDemand(2,1096,24), \textbf{D} = PhalangesOutlinesCorrect(2,2658,80), \textbf{E} = Plane(7,210,144), \textbf{F} = ShapeletSim(2,200,500), \textbf{G} = SyntheticControl(6,600,60) and \textbf{H} = ASL(10,50,53).}
\label{acc-results}
\end{table*}



\section{Evaluation}
\label{evaluation}

\subsection{Setup}
\textit{Parameter initialization} -- Because unsupervized clustering does not allow for determining the network's optimal hyperparameters through cross-validation, we use typical default parameters and ignore any dataset-specific tuning. We set the filter sizes to 16, 32 and 64 for the first, second and third convolution layers, respectively. The kernel size is set to 4x4 across all layers, the pooling size is set to 5x5 and the latent space dimensions to 10. Finally, we use L-RELU activation functions for each layer with gradient $\alpha$ = 0.1, the Adam optimizer with batch size = 256 and learning rate = 0.001, and train for 200 epochs. Our implementation is in Keras. 

\textit{Accuracy measure} -- We use the labels in the datasets to compute the accuracy of the clustering, as in DEC. Essentially, the metric finds the best matching between a ground truth label and a cluster assignment. An optimal mapping can be done with the Hungarian algorithm~\cite{hungarian}.

\textit{Datasets} -- We evaluate the performance of our approach on a collection of time series datasets from various domains, taken from the \textit{UCR Time Series Classification Archive}~\cite{ucr}. In addition, we also use the Australian Sign Language (ASL) dataset~\cite{asl}, which was used in~\cite{edr} to test the EDR distance metric. Like the authors of Autowarp~\cite{autowarp}, we use a subset consisting of $N$ = 50 time series of length $T$ = 53 from 10 different classes of signals. As opposed to the UCR datasets, which are univariate, ASL is multivariate, as measurements are provided as (x,y,z) coordinates along with the rotation of the palm.

\subsection{Accuracy}\label{acc}

We compare against two baselines that reflect the main observations previously discussed:

\begin{enumerate}
    \item \texttt{OS} (Original Space) -- classic clustering algorithms, such as k-means with Euclidean (\texttt{k-means+EUC}), k-means with DTW (\texttt{k-means+DTW}), k-shape (\texttt{k-shape}), spectral (\texttt{spectral}) or DBSCAN (\texttt{dbscan}) are applied directly on the input time series data;
    \item \texttt{LS} (Latent Space) -- raw time series are collapsed into a 5-dimension latent representation via a dense denoising AE (~\cite{dec}) and then fed into \texttt{k-means+EUC}, \texttt{spectral} or \texttt{dbscan} (i.e., note that applying \texttt{k-means+DTW} and \texttt{k-shape} on the latent representation would bring no benefits, since the obtained embeddings are in Euclidean space).
\end{enumerate}

An intuitive approach is to project the time series into an affine invariant space and then apply the same clustering algorithms as for \texttt{LS} on the transformed space. We denote it as \texttt{Pr} and also provide results for this approach. Finally, our methodology, \texttt{Pr+LS}, encompasses both affine invariant transformations and the use of a CNN-GRU AE, as described in Fig.~\ref{architecture}. We use SBD in generating the projected space. The expectation is that by using projections with an AE that can extract temporal waveforms from the original time series, we will obtain superior performance relative to the baselines. Accuracy and standard deviation over 10 runs are shown in Table~\ref{acc-results} for 7 UCR datasets and the ASL dataset. We also report improvements of \texttt{Pr+LS} and \texttt{Pr} over the \texttt{OS} and \texttt{LS} baselines (i.e., $acc_{Pr+LS|Pr} - max(acc_{OS}, acc_{LS})$) relative to the \texttt{k-means} and \texttt{spectral} algorithms respectively. We make the following observations: 
\begin{enumerate}
    \item our approach consistently outperforms both baselines and \texttt{Pr}, either with \texttt{k-means+EUC} or \texttt{spectral} -- this is best seen for dataset \texttt{H} (ASL), where accuracy reaches 82\% for \texttt{Pr+LS} with \texttt{spectral}, while the baseline models are around low-40s to mid-60s;
    \item improvements are more significant relative to \texttt{OS} and \texttt{LS}, and lower compared to \texttt{Pr}, which matches our initial intuition;
    \item improvements over \texttt{Pr} are always positive, thus justifying the need of a multi-layer CNN-GRU AE in addition to generating transformations;
    \item \texttt{Pr+LS} achieves the lowest standard deviation in the majority of cases (i.e., 1-3\% for datasets \texttt{A}-\texttt{F} and \texttt{H}), which in turn shows a high degree of robustness;
    \item mostly, \texttt{Pr+LS} with \texttt{k-means+EUC} outperforms all others models (up to 14\% improvement over \texttt{Pr+LS} with \texttt{spectral});
    \item \texttt{dbscan} is significantly less stable than \texttt{spectral} and \texttt{k-means+EUC} and highly depends on the characteristics of the underlying dataset. For most datasets its improvement with \texttt{Pr+LS} is modest over the baselines or \texttt{Pr} (i.e., 5-7\%), except for dataset \texttt{A} where it underperforms compared to \texttt{OS} by 9\% and datasets \texttt{E} and \texttt{H} where it outperforms the baselines by 69\% and 41\%, respectively. For this reason, we do no include the improvement relative to \texttt{dbscan} in Table~\ref{acc-results};
    \item \texttt{dbscan} outperforms \texttt{spectral} and \texttt{k-means+EUC} for \texttt{Pr} for datasets \texttt{B} and \texttt{D}, which is to be expected given that this algorithm is particularly suited for outlier detection (i.e., applicable for both datasets). However, this advantage does not apply when projections are used in combination with our CNN-GRU AE;
    \item with the baselines, \texttt{k-shape} is outperformed by either \texttt{k-means} or \texttt{spectral}, therefore we do not consider it in computing the improvements.
\end{enumerate}

To shed light on what happens across these datasets, we briefly discuss some of their properties. On the one hand, dataset \texttt{D} (PhalangesOutlinesCorrect) is often used for anomaly detection of hand and bone outlines and has two clusters. Both clusters' distributions show identical anomalies towards the tail of the time series. This is why the baseline models perform only slightly better than random (i.e., accuracies between 50\% and 55\%). However, by projecting the original time series and generating an effective latent representation, we are able to boost performance to 66.7\%, obtaining an improvement of 11.5\% over the best performing \texttt{k-means+EUC} model. Similar behavior of the underlying distributions is observed for datasets \texttt{C} (ItalyPowerDemand) and \texttt{F} (ShapeletSim), without any presence of anomalies. In these cases, our approach achieves significant improvements of 14\% and 22.5\%, respectively. On the other hand, dataset \texttt{E} (Plane) shows different shapes of the time series belonging to different clusters, which explains why already with most of the baseline models we achieve over 80\% accuracy. At the same time, \texttt{Pr+LS} with \texttt{k-means+EUC} and \texttt{spectral} reach 94\% accuracy. In the case of \texttt{dbscan}, accuracy is only 18\% when applied on the original time series, whereas using \texttt{LS}, \texttt{Pr} or \texttt{Pr+LS} boosts accuracy by even 69\%. This is due to the fact that \texttt{dbscan} can only handle datasets with single density~\cite{dbscan-reason}.

Finally, datasets \texttt{A} (Computers), \texttt{B} (ECG5000), \texttt{G} (SyntheticControl) and \texttt{H} (ASL) are exponents of similar underlying patterns -- some of the clusters are easily distinguishable from one another, while others are extremely similar. For example, dataset \texttt{G} contains six different classes of control charts (e.g., increasing/decreasing trend, upward/downward shift). This dataset showcases an interesting property of our approach. Some clusters could be obtained from others via geometric operations on the feature space, like rotations, so one would expect our invariant models to perform worse. However, our models are invariant to these operations \textbf{on the whole space} (not on a sample-by-sample basis), and outperform the baselines in all cases. 

\begin{figure}[h]
\centering
\begin{subfigure}[b]{.325\linewidth}
\includegraphics[width=0.98\linewidth]{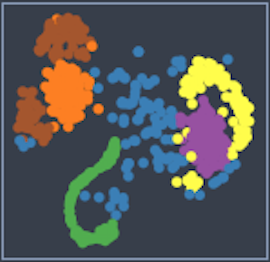}
\caption{\texttt{LS}}\label{fig:vanilla}
\end{subfigure}
\begin{subfigure}[b]{.325\linewidth}
\includegraphics[width=0.98\linewidth]{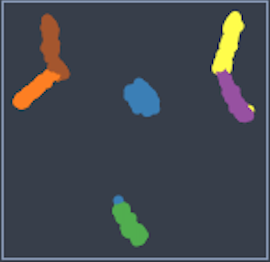}
\caption{\texttt{Pr+LS (SBD)}}\label{fig:dtw}
\end{subfigure}
\begin{subfigure}[b]{.325\linewidth}
\includegraphics[width=0.98\linewidth]{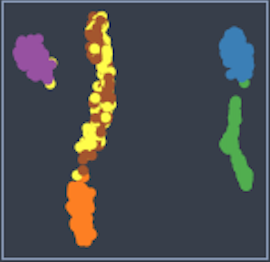}
\caption{\texttt{Pr+LS (DTW)}}\label{fig:sbd}
\end{subfigure}
\vspace{-2ex}
\caption{Latent representations after 200 epochs for the \texttt{LS} and \texttt{Pr+LS} models, when \texttt{k-means+EUC} is used as a final clustering step. For \texttt{Pr+LS}, we compare the resulting embeddings when \texttt{DTW} and \texttt{SBD} are used.} 
\label{fig:latent}
\end{figure}

In Fig.~\ref{fig:latent} we compare the latent representation generated by the \texttt{LS} model (i.e., recall that the AE used here is a denoising dense AE, as defined in DEC), with those obtained with \texttt{Pr+LS}, when either DTW and SBD are used as metrics in computing the projections. In all three cases, we apply \texttt{k-means+EUC} as the final clustering step. An accuracy of 84\% is achieved by using SBD (as reported in Table~\ref{acc-results}), whereas with \texttt{LS} performance is only at 57\%. These results are explained by how well separated clusters are with SBD (Fig.~\ref{fig:dtw}) compared to a simple AE (Fig.~\ref{fig:vanilla}), and even to using DTW instead (Fig.~\ref{fig:sbd}), which performs 12\% worse.

\subsection{Sensitivity to number of pivot points}\label{pivots}

\begin{figure}[t]
  \includegraphics[width=1.00\linewidth]{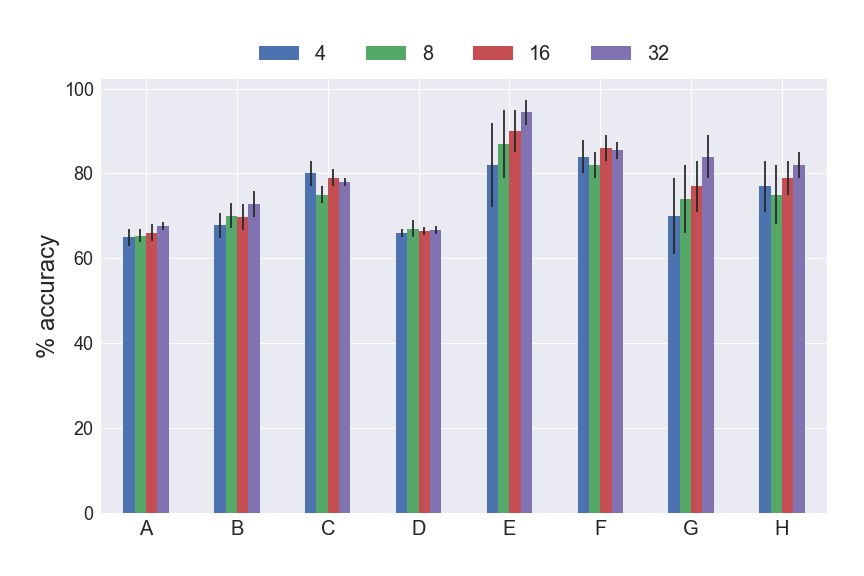}
  \vspace{-5ex}
  \caption{Accuracy and standard deviation for datasets A-H when $p = \{4,8,16,32\}$ pivot points are used to generate the projected space.} 
  \label{pivotpoints}
\end{figure}

To generate the projected space for \texttt{Pr+LS}, we pick $p$ random \textit{pivot points} from the initial dataset and represent every sample as its similarity to each pivot point. We show accuracy and standard deviation results for $p$ $\in$ \{4,8,16,32\}, when SBD is used, in Fig.~\ref{pivotpoints}. We omit results for DTW, since they show similar trends. In terms of accuracy, on the one hand, we note that for datasets \texttt{A}, \texttt{B}, \texttt{C}, \texttt{D} and \texttt{F}, differences in accuracy are marginal (2\%, 5\%, 2\%, 1\%, 4\%), considering the standard deviation as well. On the other hand, for datasets \texttt{E}, \texttt{G} and \texttt{H} accuracy increases steadily with $p$, by 12.4\%, 15.5\% and 7\% respectively. Across all datasets standard deviation decreases as $p$ increases which - coupled with increased (or relatively constant) accuracy - shows that the preferred number of pivot points is 16 or 32. Naturally, $p$ can take values greater than 32. We evaluated the evolution of accuracy and standard deviation in such cases but noticed that while standard deviation reduces \textit{slightly}, the accuracy does not improve any further.

One possible improvement to randomly selecting pivot points would be to leverage optimal stopping theory~\cite{bruss2003note}. In that scenario we would introduce a slight overhead in picking optimal pivots but we would only need to have $p \approx k$ (as many pivots as the number of clusters), lowering overall temporal complexity while maintaining low standard deviation and high accuracy.

\subsection{Computation cost}

A secondary advantage of our approach, apart from the superior accuracy, is its low overhead through the optimization previously described (i.e., applying transformations directly on the input space as a pre-processing step). Given this, the overall added complexity is $\mathcal{O}\left(N \cdot p \cdot C\right)$, where $N$ is the number of samples, $p$ is the number of pivot points and $C$ is the complexity of the selected similarity metric. We note the following: (i) $p$ will be a constant with a value much lower than $N$, as previously shown in the sensitivity analysis, and (ii) $C$ can be relatively low as the algorithm is not dependent on the selected similarity metric, as reflected in the accuracy results. These observations lead to a complexity growth that is linear on the number of samples, namely $\mathcal{O}\left(N \cdot C\right)$. 

We measure the time required to generate the transformations with either DTW or SBD as the similarity metric, when $p$ = 16 pivot points are used. For instance, for dataset $C$, both times are comparable (~1s), while for datasets $D$, $E$, $F$ and $G$, using DTW incurs execution times 3 to 32 times higher than in the case of SBD (i.e., (32s, 5s, 67s, 3s) compared to (1s, 1s, 5s, 1s)). All experiments were run on CPUs only, without any dedicated hardware (either GPU or onboard graphics). Each of the CPUs has 12 virtual cores clocked at \textasciitilde2.33GHz.

\section{Conclusions}

In this paper, we tackle the problem of clustering temporal data. While a plethora of traditional, time series specific and more recent deep learning-based clustering methods exist, they suffer from important drawbacks when both varying sampling rates and high dimensionality (i.e., long sequences) are present in the input data. 

Therefore, we propose a novel approach in which we first transform the time series into a distance-based projected representation by using elastic metrics (DTW) or cross-correlation measures (SBD) suitable for dealing with temporal data. Then, we feed these projections into a multi-layer CNN-GRU AE in order to generate meaningful latent representations, that allow for a natural separation of clusters with any Euclidean distance-based clustering algorithm. Evaluation on multiple univariate and multivariate time series datasets from various domains shows that our approach is robust and outperforms existing methods in all cases: up to 32\% for spectral clustering, and up to 22.5\% for k-means clustering. The computational overhead of our method is negligible.


\newpage

\begin{quote}
\begin{small}
\bibliographystyle{aaai}
\bibliography{main}
\end{small}
\end{quote}

\end{document}